\documentclass{article} 
\usepackage[accepted]{icml2020}

\usepackage[algo2e]{algorithm2e} 

\usepackage{hyperref}
\usepackage{url}
\usepackage{graphicx}
\usepackage{amsmath}
\usepackage{bbm}
\usepackage{multirow}

\newcommand{\mb}{\mathbf}
\newcommand{\mc}{\mathcal}

\newtheorem{definition}{\textsc{Definition}}
\newtheorem{theorem}{\textsc{Theorem}}

\newcommand{\our}{\textsc{GB-Distance}}
\newcommand{\gbert}{\textsc{Graph-Bert}}
\newcommand{\bert}{\textsc{Bert}}

\begin{document}

\twocolumn[
\icmltitle{Graph Neural Distance Metric Learning with {\gbert}}



\icmlsetsymbol{equal}{*}

\begin{icmlauthorlist}
\icmlauthor{Jiawei Zhang}{ifmlab}
\end{icmlauthorlist}

\icmlaffiliation{ifmlab}{IFM Lab, Department of Computation, Florida State University, Tallahassee, FL, USA.}

\icmlcorrespondingauthor{Jiawei Zhang}{jiawei@ifmlab.org}

\icmlkeywords{Machine Learning, ICML}

\vskip 0.3in
]

\printAffiliationsAndNotice{}

\begin{abstract}

Graph distance metric learning serves as the foundation for many graph learning problems, e.g., graph clustering, graph classification and graph matching. Existing research works on graph distance metric (or graph kernels) learning fail to maintain the basic properties of such metrics, e.g., \textit{non-negative}, \textit{identity of indiscernibles}, \textit{symmetry} and \textit{triangle inequality}, respectively. In this paper, we will introduce a new graph neural network based distance metric learning approaches, namely {\our} ({\gbert} based Neural Distance). Solely based on the attention mechanism, {\our} can learn graph instance representations effectively based on a pre-trained {\gbert} model. Different from the existing supervised/unsupervised metrics, {\our} can be learned effectively in a semi-supervised manner. In addition, {\our} can also maintain the distance metric basic properties mentioned above. Extensive experiments have been done on several benchmark graph datasets, and the results demonstrate that {\our} can out-perform the existing baseline methods, especially the recent graph neural network model based graph metrics, with a significant gap in computing the graph distance.

\end{abstract}
\section{Introduction}\label{sec:introduction}

Graph provides a general representation of many network structured data instances in the real world, which can capture both the properties of nodes and the extensive connections among the nodes. For instance, the app function-call diagrams \cite{HCS09}, brain-region functional activities \cite{BB11} and the bio-medical drug molecules \cite{DMIBHAA15} can all be represented as graphs in various shapes. An important research problem in graph studies is to learn the distance metric of the graph instances \cite{B97, GXTL10, BS98}, which can serve as the foundation of many other research tasks, e.g., molecular graph clustering \cite{JBJ18}, brain graph classification \cite{RABV13} and frequent sub-graph extraction \cite{YH02}. In this paper, we will not distinguish the differences among graph distance, graph similarity and graph kernel learning problems, and unify them all as the graph distance metric learning problem.

Graph instance distance metric learning is an interesting research problem, and many research works have been done on this topic. In early years, to measure the distance between graphs, graph edit distance \cite{B97, GXTL10} and maximum common subgraph \cite{BS98} are commonly used. The distance metrics defined by these two methods can convey concrete physical meanings (i.e., edit distances and subgraph), but these methods are also known to be NP-complete \cite{BS98, Zeng_Comparing_09}. In the past few years, we have witnessed new developments on graph distance metric learning, many of them are based on the graph neural network models \cite{LGDVK19, Bai_SimGNN_19}. In \cite{LGDVK19}, a new model named graph matching network is introduced, which adopts the propagation layer in the model for learning the node distance scores between graphs. Meanwhile, in \cite{Bai_SimGNN_19}, the authors propose to apply the latest graph convolutional network for node representation learning prior to computing the graph distance scores. 

However, via a thorough analysis about the existing graph distance metrics, several common disadvantages about them can be identified, which are listed as follows:
\begin{itemize}

\item \textbf{High Computational Cost}: For the traditional graph edit distance or subgraph learning based methods, the metric score computational process can be extremely time-consuming. Meanwhile, for the pair-wise graph neural distance metrics, the model training cost will grow quadratically as the graph number increases.

\item \textbf{Node-Order Invariant Representation}: For the latest graph neural network based methods, which take neural network models as the representation learning component, the learned graph instance representations and the distance metric scores will vary greatly as the input graph node order changes. 

\item \textbf{Semi-Supervised Learning}: To train the neural network based methods, a large number of graph-pair distance scores need to be labeled in advance, which can be very tedious and time consuming. The existing works mostly fail to utilize the unlabeled graph pairs in the metric learning.

\item \textbf{Lack of Metric Properties}: Furthermore, for most of the existing graph distance metric learning approaches, they fail to maintain the basic properties of the metrics \cite{STD05, metric_tutorial} in the learning process, like \textit{non-negativity}, \textit{identity of indiscernibles}, \textit{symmetry} and \textit{triangle inequality}, respectively.
\end{itemize}

In this paper, we aim to introduce a new graph distance metric learning approach, namely {\our} ({\gbert} based Neural Distance), to resolve the disadvantages with the existing works mentioned above. {\our} is based on the state-of-the-art {\gbert} model \cite{zhang2020graphbert}, which is capable to learn effective graph representations based on the attention mechanism. 

Meanwhile, for efficient and effective metric scores computation, {\our} further modifies {\gbert} in several major perspectives by (1) extending {\gbert} for graph instance representation learning ({\gbert} is proposed for graph node embedding originally), (2) introducing pre-training and fine-tuning to graph neural distance metric learning to lower down the learning costs, and (3) proposing new node-order invariant initial input embeddings and model functional components. What's more, {\our} works very well in the semi-supervised learning setting and can also effectively incorporate various metric properties in the learning process as additional constraints of the objective function. 

The remaining sections of this paper are organized as follows. We will first talk about the related works in Section~\ref{sec:related_work}, and then introduce the notations, terminology definitions and problem formulation in Section~\ref{sec:formulation}. Detailed information of the {\our} model is provided in Section~\ref{sec:method}, whose effectiveness will be tested with experiments on real-world benchmark datasets in Section~\ref{sec:experiment}. At the end, we will conclude this paper in Section~\ref{sec:conclusion}.

\section{Related Work}\label{sec:related_work}

In this section, we will briefly introduce the related work on \textit{graph neural network}, \textit{graph metric learning}, \textit{metric optimization} and \textit{BERT}.

\noindent \textbf{Graph Neural Network}: In addition to the graph convolutional neural network \cite{Kipf_Semi_CORR_16} and its derived variants \cite{Velickovic_Graph_ICLR_18,Li_Deeper_CORR_18,sun2019adagcn,DBLP:journals/corr/abs-1907-02586}, many great research works on graph neural networks have been witnessed in recent years. In \cite{Meng_Isomorphic_19}, the authors introduce the graph isomorphic neural network, which can automatically learn the subgraph patterns based representations for graphs. In \cite{zhang2020graphbert}, the authors introduce a new type of graph neural network based on graph transformer and BERT. Many existing graph neural network models will suffer from performance problems with deep architectures. In \cite{Zhang2019GResNetGR,Li_Deeper_CORR_18,sun2019adagcn,Huang_Inductive_19}, the authors explore to build deep graph neural networks with residual learning, dilated convolutions, and recurrent network, respectively. A comprehensive survey of existing graph neural networks is also provided in \cite{zhang2019graph,DBLP:journals/corr/abs-1901-00596}.

\noindent \textbf{Graph Metric Learning}: In addition to the classic graph edit distance \cite{B97, GXTL10} and common subgraph \cite{BS98} based graph distance metrics, there also exist several important recent research works on graph distance metric learning \cite{kondor2016multiscale,DBLP:journals/corr/abs-1906-01277,Yanardag_Deep_15,LGDVK19,Bai_SimGNN_19,pmlr-v5-shervashidze09a}. \cite{pmlr-v5-shervashidze09a} proposes to compare graphs by counting graphlets; \cite{Yanardag_Deep_15} learns latent representations of sub-structures for graphs with deep neural networks; \cite{DBLP:journals/corr/abs-1906-01277} defines the graph kernels based on the Wasserstein distance between node feature vector distributions of two graphs; and \cite{kondor2016multiscale} introduces a multi-scale Laplacian graph kernel instead. \cite{LGDVK19} adopts the propagation layer in the model for learning the node distance scores between graphs, and \cite{Bai_SimGNN_19} utilizes the graph convolutional network for representation learning prior to computing graph distance. However, these existing graph metric learning methods suffer from either the lack of interpretability or the high computational costs.

\noindent \textbf{Metric Optimization}: Distance metric learning is a classic research task in machine learning, a comprehensive survey on which is also available at \cite{yang2006distance}. In \cite{NIPS2002_2164}, the authors formulate the distance metric learning as a convex optimization problem, which also proposes an efficient and local-optima-free learning algorithm to solve the problem. In \cite{10.5555/2188385.2188386}, the distance metric learning is defined as an eigenvalue optimization problem instead. In \cite{STD05}, the authors further consider the basic properties on distance metrics, which formulate the problem as a constrained optimization problem. They also introduce an efficient learning algorithm to maintain the triangle inequality subject to the $L_p$ norm.

\noindent \textbf{BERT and Transformer}: In NLP, the dominant sequence transduction models are based on complex recurrent \cite{Hochreiter_Long_Neural_97,DBLP:journals/corr/ChungGCB14} or convolutional neural networks \cite{kim-2014-convolutional}. However, the inherently sequential nature precludes parallelization within training examples. Therefore, in \cite{VSPUJGKP17}, the authors propose a new network architecture, i.e., the Transformer, based solely on attention mechanisms, dispensing with recurrence and convolutions entirely. With Transformer, \cite{DCLT18} further introduces {\bert} for deep language understanding, which obtains new state-of-the-art results on majority of the natural language processing tasks.


\section{Notations and Problem Formulation}\label{sec:formulation}

In this section, we will introduce the notations used in this paper, and provide the formulation of the studied problem.

\subsection{Notations}

In the sequel of this paper, we will use the lower case letters (e.g., $x$) to represent scalars, lower case bold letters (e.g., $\mb{x}$) to denote column vectors, bold-face upper case letters (e.g., $\mb{X}$) to denote matrices, and upper case calligraphic letters (e.g., $\mathcal{X}$) to denote sets or high-order tensors. Given a matrix $\mb{X}$, we denote $\mb{X}(i,:)$ and $\mb{X}(:,j)$ as its $i_{th}$ row and $j_{th}$ column, respectively. The ($i_{th}$, $j_{th}$) entry of matrix $\mb{X}$ can be denoted as either $\mb{X}(i,j)$ or $\mb{X}_{i,j}$, which will be used interchangeably. We use $\mb{X}^\top$ and $\mb{x}^\top$ to represent the transpose of matrix $\mb{X}$ and vector $\mb{x}$. For vector $\mb{x}$, we represent its $L_p$-norm as $\left\| \mb{x} \right\|_p = (\sum_i |\mb{x}(i)|^p)^{\frac{1}{p}}$. The Frobenius-norm of matrix $\mb{X}$ is represented as $\left\| \mb{X} \right\|_F = (\sum_{i,j} |\mb{X}(i,j)|^2)^{\frac{1}{2}}$. The element-wise product of vectors $\mb{x}$ and $\mb{y}$ of the same dimension is represented as $\mb{x} \otimes \mb{y}$, whose concatenation is represented as $\mb{x} \sqcup \mb{y}$.

\subsection{Problem Formulation}

The data instances studied in this paper are all in the graph structure, which can be denoted as the graph instance.
\begin{definition}
(Graph Instance): Formally, a graph instance can be represented as $G = (\mc{V}, \mc{E}, w, x)$, where $\mc{V}$ and $\mc{E}$ denote the sets of nodes and links, respectively. Mapping $x: \mc{V} \to \mc{X}$ projects the nodes to their corresponding raw attributes in space $\mc{X}$. For presentation simplicity, we can also denote the raw feature vector of $v_i \in \mc{V}$ as $\mb{x}_i = x(v_i) \in \mc{X}$. Meanwhile, the mapping $w: \mc{V} \times \mc{V} \to \mathbbm{R}$ can project node pairs to their corresponding link weights. For any non-existing link $(v_i, v_j) \in \mc{V} \times \mc{V} \setminus \mc{E}$, we have $w(v_i, v_j) = 0$ by default.
\end{definition}

The above definition provides a general representation for graph instances. For the graph raw attributes, they can denote various types of information actually depending on the application settings, e.g., images, textual descriptions and simple tags. Meanwhile, if the graph instances studied are unweighted, we will have $w(v_i, v_j) = 1, \forall (v_i, v_j) \in \mc{R}$, and $w(v_i, v_j) = 0, \forall (v_i, v_j) \in \mc{V} \times \mc{V} \setminus \mc{E}$. Based on the above definition, we can define the problem studied in this paper as follows.

\noindent \textbf{Problem Statement}: Formally, given a set of $m$ graph instances $\mc{G} = \{G^{(1)}, G^{(2)}, \cdots, G^{(m)}\}$ (the superscript denotes the graph index), in this paper, we aim to learn a mapping $d: \mc{G} \times \mc{G} \to \mathbbm{R}$ to compute the distance for any graph instance pairs from $\mc{G}$. Here, mapping $d(\cdot, \cdot)$ should also maintain the basic distance metric properties, i.e., \textit{non-negativity}, \textit{identity of indiscernibles}, \textit{symmetry} and \textit{triangle inequality}. Detailed representation of these different properties will be illustrated in the following section.
\section{Method}\label{sec:method}

In this section, we will introduce the {\our} model. At the beginning, we will provide the description of the architecture of {\our} first, whose internal functional components will be introduced in detail in the follow-up subsections. At the end, we will talk about the mathematical constraints for modeling the metric properties in the learning process.

\subsection{Framework Description}\label{subsec:outline}

\begin{figure*}[t]
    \begin{minipage}{\textwidth}
    \centering
    	\includegraphics[width=0.85\linewidth]{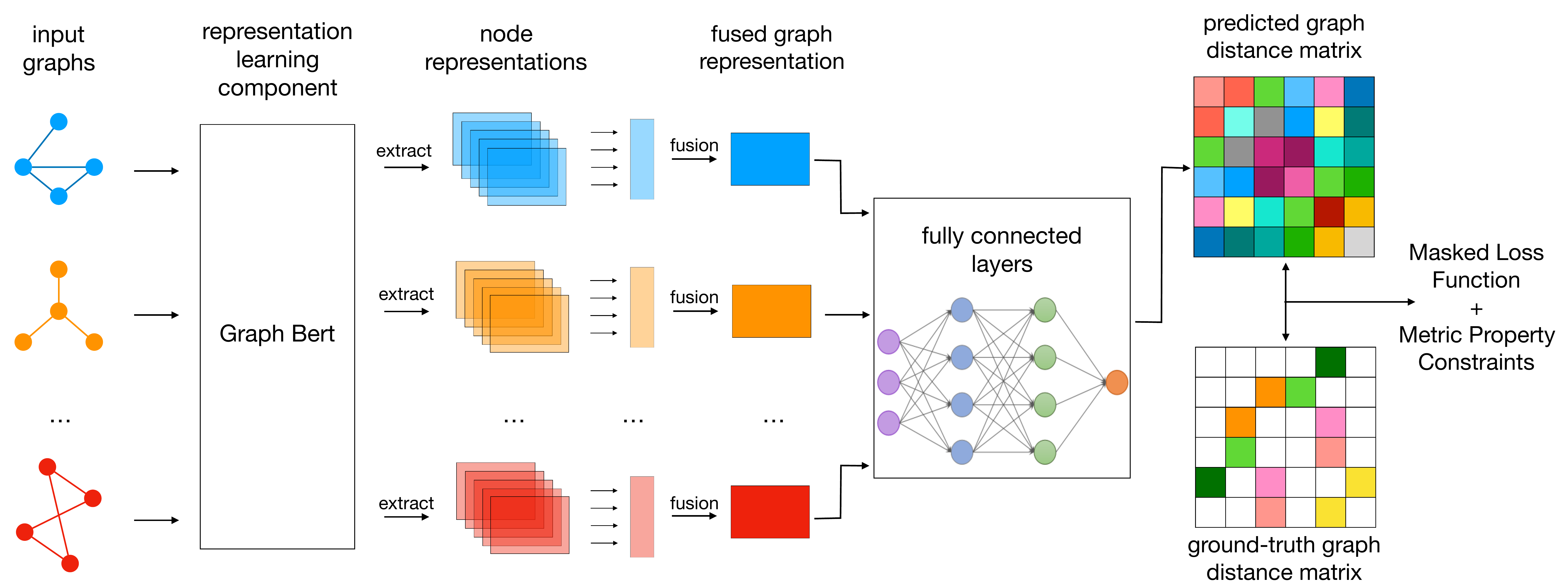}
    	\caption{An Illustration of the {\our} Framework for Graph Distance Metric Learning.}
    	\label{fig:architecture}
    \end{minipage}%
\end{figure*}

As illustrated in Figure~\ref{fig:architecture}, given a set of graph instances, e.g., $\mc{G} = \{G^{(1)}, G^{(2)}, \cdots, G^{(m)} \}$, {\our} can effective compute the pairwise distance scores among them with several key functional components:
\begin{itemize}
\item \textit{{\gbert} Layers}: To effectively and efficiently extract the feature representations of the input graph instances, {\our} proposes to extend the {\gbert} model \cite{zhang2020graphbert} to the graph instance representation learning settings. Different from the existing graph neural networks, {\gbert} learns node representations merely based on the attention mechanisms, which will not suffer from the common performance problems with the existing graph neural networks. Also {\our} can be pre-trained in an unsupervised manner, which can greatly lower down the time costs for model learning.

\item \textit{Representation Fusion}: In addition, different from the target node representation learning \cite{zhang2020graphbert}, in this paper, we focus on learning the representations of graph instances instead. In {\our}, we introduce a fusion component to integrate the learned node representations as the representation of the whole graph instances.

\item \textit{Distance Metric Inference}: {\our} proposes to apply fully connected layers to compute the distance metrics between pairwise graph instances. Prior to feeding the learned graph instance representations, necessary representation vector comparison operators will also be needed, which is not illustrated in the plot.

\item \textit{Masked Loss Function}: {\our} doesn't require a large number of labeled graph pair distance scores as the training data, which can also effectively involve the unlabeled graph pairs in defining the loss function with a mask matrix. Formally, in the masked loss function, the labeled and unlabeled graph pairs will be assigned with different weights.

\item \textit{Metric Property based Constraints}: The graph distance metric properties, including \textit{non-negativity}, \textit{identity of indiscernibles}, \textit{symmetry} and \textit{triangle inequality}, will introduce different mathematical constraints (or involved as a functional components) to define the final objective function. {\our} will learn both the model variables, graph representations, and their pairwise distance metric scores by optimizing the objective function. Necessary model pre-training will be adopted in {\our} as well so as to lower down the learning time costs. Furthermore, to handle the triangle inequality constraints, we will introduce a post-process algorithm to resolve the violations in the learned results.
\end{itemize}

Detailed information about these five functional components in {\our} will be introduced in detail in the following subsections.

\subsection{{\gbert} based Layers and Pre-Training}\label{subsec:graph_bert}

In this part, we will talk about the graph representation learning layer used in {\our}, which can compute the node-order-invariant graph representations effectively. Formally, given an input graph instance $G \in \mc{G}$, we can denote its node set as $\mc{V}$ (here, we will not indicate the graph instance index for representation simplicity). The relative node positions in the list will not change the nodes' learned representations in {\our}. Therefore, for presentation simplicity, regardless of the node orders, we can also serialize the nodes into a list as $[v_1, v_2, \cdots, v_{|\mc{V}|}]$.

For each node in the list, e.g., $v_i$, we can represent its raw features as a vector $\mb{x}_i = x(v_i) \in \mathbbm{R}^{d_x \times 1}$ as defined in Section~\ref{sec:formulation}, which may cover various types of information, e.g., node tags, attributes, textual descriptions and even images. Via certain embedding functions, we can denote the embedded feature representation of $v_i$'s raw features as
\begin{equation}
\mb{e}_i^{x} = \mbox{Embed} \left( \mb{x}_i \right).
\end{equation}
Here, the embedded feature vector $\mb{e}_i^{x} \in \mathbbm{R}^{d_h \times 1}$ and $d_h$ denotes its vector length. Meanwhile, depending on the input features, different approaches can be utilized to define the $\mbox{Embed}(\cdot)$ function, e.g., CNN for image features, LSTM for textual features, positional embedding for tags and MLP for real-number features.

In addition to the node raw feature embedding, we also define the nodes' Weisfeiler-Lehman role embedding vector in this paper, which effectively denotes the nodes' global roles in the input graph. As introduced in \cite{zhang2020graphbert}, nodes' Weisfeiler-Lehman code is node-order-invariant, which denotes a positional property of the nodes actually. Formally, given a node $v_i$ in the input graph instance, we can denote its pre-computed WL code as $\mbox{WL}(v_i) \in \mathbbm{N}$, whose corresponding embeddings can be represented as
\begin{equation}
\begin{aligned}
\mb{e}_i^{r}  &= \mbox{Position-Embed}\left( \mbox{WL}(v_i) \right)\\
&= \left[sin\left (\frac{\mbox{WL}(v_i)}{10000^{\frac{2 l}{d_{h}}}} \right), cos\left(\frac{\mbox{WL}(v_i)}{10000^{\frac{2 l + 1}{d_{h}}}} \right) \right]_{l=0}^{\left \lfloor \frac{d_h}{2} \right \rfloor},
\end{aligned}
\end{equation}
where $\mb{e}_i^{r} \in \mathbbm{R}^{d_h \times 1}$ and $l$ denotes the vector index $l$.

Both the node raw attribute embedding and WL role embedding are inherit from \cite{zhang2020graphbert}. Meanwhile, to handle the graph instances without node attributes, we also introduce two new embeddings based on the nodes' fix-order neighborhood and nodes' degrees, which can be denoted as follows
\begin{equation}
\begin{aligned}
\mb{e}_i^{w} &= \mbox{Embed} \left( \mb{w}_i \right) \in \mathbbm{R}^{d_h \times 1},\\
\mb{e}_i^{d}  &= \mbox{Position-Embed}\left( \mbox{D}(v_i) \right) \in \mathbbm{R}^{d_h \times 1},
\end{aligned}
\end{equation}
where $\mb{w}_i = [w(v_i, v_j)]_{v_j \in \mc{V}} \in \mathbbm{R}^{|\mc{V}| \times 1}$ denotes the connection weights between $v_i$ and the other nodes in the graph and $D(v_i) \in \mathbbm{N}$ is the degree of $v_i$. To ensure vector $\mb{w}_i$ is node-order invariant, we cast an artificial fixed node order for the vector entries. In addition, for all the nodes in the identical graph, such an artificial node order will be the same. 

By aggregating the above four embedding vectors together, we can define the initial input embeddings for node $v_i$ as
\begin{equation}\label{equ:initial_embedding}
\mb{h}_i^{(0)} = \mbox{sum} \left( \mb{e}_i^{(x)}, \mb{e}_i^{(r)}, \mb{e}_i^{w}, \mb{e}_i^{d}\right) \in \mathbbm{R}^{d_h \times 1}.
\end{equation}

Among all the graphs in $\mc{G}$, we can represent the largest graph instance size as $k_{max}$. Furthermore, the initial embedding vectors of all the nodes in graph $G$ can be organized as a matrix $\mb{H}^{(0)} = [\mb{h}_1^{(0)}, \mb{h}_2^{(0)}, \cdots, \mb{h}_{|k_{max}|}^{(0)}]^\top \in \mathbbm{R}^{k_{max} \times d_h}$. For the graph instances with less than $k_{max}$ nodes, we will adopt zero padding to expand the matrix to the dimensions specified above.

{\our} uses the graph-transformer based encoder to update the nodes' representations iteratively with multiple layers ($D$ layers) iteratively as follows:
\begin{equation}
\begin{cases}
\vspace{5pt}
\mb{H}^{(0)} & \hspace{-10pt} = [\mb{h}_1^{(0)}, \mb{h}_{2}^{(0)}, \cdots, \mb{h}_{|\mc{V}|}^{(0)}]^\top ,\\
\vspace{5pt}
\mb{H}^{(l)} &\hspace{-10pt}= \mbox{G-Transformer} \left( \mb{H}^{(l-1)}\right), \forall l \in \{1, 2, \cdots, D\},\\
\mb{z} &\hspace{-10pt}= \mbox{Fusion} \left( \mb{H}^{(D)} \right). 
\end{cases}
\end{equation}
Operator $\mbox{G-Transformer}(\cdot)$ denotes the graph-transformer, whose concrete representation can be illustrated as follows:
\begin{equation}
\begin{aligned}
\hspace{-7pt} \mb{H}^{(l)} &= \mbox{G-Transformer} \left( \mb{H}^{(l-1)}\right)\\
&= \mbox{softmax} \left(\frac{\mb{Q} \mb{K}^\top}{\sqrt{d_h}} \right) \mb{V} + \mbox{G-Res} \left( \mb{H}^{(l-1)}, \mb{X}_i\right),
\end{aligned}\vspace{-2pt}
\end{equation}
where
\begin{equation}
\begin{cases}
\mb{Q} & = \mb{H}^{(l-1)} \mb{W}_Q^{(l)},\\
\mb{K} & = \mb{H}^{(l-1)} \mb{W}_K^{(l)},\\
\mb{V} & = \mb{H}^{(l-1)} \mb{W}_V^{(l)}.\\
\end{cases}\vspace{-2pt}
\end{equation}
In the above equations, $\mb{W}_Q^{(l)}, \mb{W}_K^{(l)}, \mb{W}_K^{(l)} \in \mathbbm{R}^{d_h \times d_h}$ are the involved variables. To simplify the presentations in the paper, we assume nodes' hidden vectors in different layers have the same length. Notation $\mbox{G-Res} \left( \mb{H}^{(l-1)}, \mb{X}_i\right)$ represents the graph residual term introduced in \cite{Zhang2019GResNetGR}. Different from the original {\gbert} in \cite{zhang2020graphbert}, which forces to add the residual terms of the target to all the nodes in the sub-graphs, the residual terms used here correspond to all the nodes in the graph instance instead. 

Furthermore, the operator $\mbox{Fusion}(\cdot)$ will aggregate such learned nodes' representations to define the representation of the graph instances. Formally, we can rewrite such an operator as follows:
\begin{equation}
\mb{z} = \mbox{Fusion} \left( \mb{H}^{(D)} \right) = \frac{1}{k_{max}} \sum_{i = 1}^{k_{max}} \mb{h}_{i}^{(D)}.
\end{equation}

In this paper, we use the simple node representation averaging to define the fusion component for graph instance representation learning. Finally, vector $\mb{z}$ will be outputted as the learned representation of the input graph instance $G$. Meanwhile, to ensure such learned representations can capture the graph information, in this paper, we also propose to pre-train the {\gbert} layer in advance with the \textit{node raw attribute reconstruction} and \textit{graph structure recovery} tasks concurrently as introduced in \cite{zhang2020graphbert}. These two pre-training task both work in an unsupervised learning manner, and the pre-training time cost is only decided by the available graph instance numbers, which is very minor compared against the graph pairwise distance metric optimization to be discussed later. The pre-training allows {\our} to initialize the graph-transformer layers with a good state, which will greatly lower down the learning cost {\our} greatly afterwards.

\subsection{Distance Metric Inference} 

Based on the above descriptions, we can denote the learned representations for all the graph instances in $\mc{G}$ as set $\left\{\mb{z}^{(i)}\right\}_{G^{(i)} \in \mc{G}}$. {\our} can effective project the graph instance pairs to their corresponding distance metric values with several fully connected (FC) layers. Formally, given an input graph pair $G^{(i)}$ and $G^{(j)}$, we can represent their fused representations as $\mb{z}^{(i)}$ and $\mb{z}^{(j)}$, respectively. The distance between them can be inferred effectively in {\our} as follows:
\begin{equation}\label{equ:distance}
\begin{aligned}
& d(G^{(i)}, G^{(j)}) = 1.0 - \exp \left( - \mbox{FC} \left( ( \mb{z}^{(i)}  -  \mb{z}^{(j)} )**2 \right) \right),
\end{aligned}
\end{equation}
where $(\cdot)**2$ denotes the entry-wise square of the input vector and $\mbox{FC}(\cdot)$ represents the fully connected layers. To ensure the learned distance metric is \textit{symmetric}, we compute $( \mb{z}^{(i)}  -  \mb{z}^{(j)} )**2$ in the model instead of simple vector concatenation. Meanwhile, for the output layer in $\mbox{FC}(\cdot)$, function $\exp^{-x}$ is adopted to ensure the learned distance metric value is \textit{non-negative} and within a normalized range $[0, 1]$. According to the above definition, it is easy to know that the \textit{identity of indiscernibles} property can be effectively maintained, as $d(G^{(i)}, G^{(i)}) = 0$ holds for $\forall G^{(i)} \in \mc{G}$.

Based on the inference model, we can represent the pairwise graph distance metric values as a matrix ${\mb{D}} \in \mathbbm{R}^{m \times m}$ ($m$ denotes the graph instance set size), where entry ${\mb{D}}(i, j) = d(G^{(i)}, G^{(j)})$. According to the above model description, we can know that matrix ${\mb{D}}$ is symmetric since $d(G^{(i)}, G^{(j)}) = d(G^{(j)}, G^{(i)})$ holds for any graph pairs. Meanwhile, based on the graph true distance metric (which will be introduced in Section~\ref{sec:experiment} in detail), for the graph pairs in the training set, we can represent them as the ground-truth matrix $\bar{\mb{D}} \in \mathbbm{R}^{m \times m}$, where the diagonal entries are assigned with value $0$ (to denote they are extremely close or identical) and entries corresponding to the unlabeled graph pairs are filled in with value $1$ by default (to denote they are far away). To effectively incorporate both the labeled and unlabeled graph pairs in the model learning, we introduce the masked loss function as follows:
\begin{equation}
\ell (\mb{D}) = \left\| \mb{M} \odot (\mb{D} - \bar{\mb{D}}) \right\|_p.
\end{equation}
where $\left\|\cdot \right\|_p$ denotes the $L_p$ matrix norm and $\mb{M} \in \mathbbm{R}^{m \times m}$ denotes the mask matrix with entry 
\begin{equation}\label{equ:mask_weight}
\mb{M}(i,j) = \begin{cases}
1, & \mbox{ if } \bar{\mb{D}}(i,j) \mbox{ is labeled;}\\
\alpha, & \mbox{ if } \bar{\mb{D}}(i,j) \mbox{ is unlabeled} \land i \neq j;\\
\beta, & \mbox{ if } i = j.
\end{cases}
\end{equation}
In the above equation, $\alpha \in [0, 1]$ is a hyper-parameter which can be fine-tuned with the validation set, and $\beta$ is usually a very large number (e.g., $10^3$) to force the \textit{identity of indiscernibles} property can hold.

\subsection{Metric Property based Constraints}

In the above model architecture introduction, we have accommodated the function components to incorporate several key properties of the distance metric, including \textit{symmetry} (in Equation~\ref{equ:distance}), \textit{non-negativity} (with the $\exp^{-x}$ function for the output layer in $FC(\cdot)$ operator), and \textit{identity of indiscernibles} (in both Equation~\ref{equ:distance} and Equation~\ref{equ:mask_weight}), respectively. Here, we will tackle the last important property on the distance metric, i.e., \textit{triangle inequality}.

Formally, for the distance metric $d(\cdot, \cdot)$, we can represent the \textit{triangle inequality} on any three graph instances $G^{(i)}, G^{(j)}, G^{(k)} \in \mc{G}$ with the following equation:
\begin{equation}
d(G^{(i)}, G^{(j)}) \le d(G^{(i)}, G^{(k)}) + d(G^{(k)}, G^{(j)}).
\end{equation}

If we represent such constraints based on the distance matrix $\mb{D}$ to be inferred, it can be denoted as
\begin{equation}
\hspace{-5pt}\mb{D}(i,j) \le \mb{D}(i,k) + \mb{D}(j,k), \forall i, j, k \in \{1, \cdots, m\}.
\end{equation}

Based on it, we can represent the overall framework objective function of the graph neural distance metric learning problem as follows:
\begin{equation}\label{equ:objective}
\hspace{-10pt}
\begin{aligned}
& \min \left\| \mb{M} \odot (\mb{D} - \bar{\mb{D}}) \right\|_p \\
& s.t. \mb{D}(i,j) \hspace{-2pt} \le \hspace{-2pt} \mb{D}(i,k) \hspace{-2pt}+\hspace{-2pt} \mb{D}(j,k), \forall i, j, k \in \{1, \cdots, m\}.
\end{aligned}
\end{equation}
where both the model variables and the distance matrix variable $\mb{D}$ are to be optimized concurrently.

\subsection{Practical Issues in Framework Learning}\label{subsec:framework}

To learn the mode, we can denote $\mc{D}$ as the collection of all potential distance matrices which can meet the constraints. As inspired by \cite{STD05}, we can prove that the objective function is learnable, and its global optimum can also be identified subject to certain conditions with the following theorem.

\begin{algorithm}[t]
	\caption{Triangle-Fixing ($\hat{\mb{D}}$, $\epsilon$)}
	\label{alg:framwork}
	\small
	\KwIn{Learned graph distance matrix $\hat{\mb{D}}$; Parameter $\epsilon$.}
	\KwOut{Inferred matrix $\mb{D} = \arg\min_{\mb{D} \in \mc{D}} \left\|\mb{D} - \hat{\mb{D}} \right\|$.}
	\Begin{	
		\For{$1 \le i < j < k \le m$}{
			Initialize variable $z_{ijk} = 0$; 
		}
		\For{$1 \le i < j \le m$}{
			Initialize variable $e_{ij} = 0$; 
		}
		$\delta = 1 + \epsilon$\\
		\While{($\delta > \epsilon$)}
		{
			\For{each violated triangle (i, j, k)}{
				{$b = \hat{\mb{D}}(k,i) + \hat{\mb{D}}(j,k) - \hat{\mb{D}}(i,j)$\\
				$\mu = -\frac{1}{3} (b - e_{ij} + e_{jk} + e_{ki})$\\
				$\theta = \min\{\mu, z_{ijk}\}$\\
				$e_{ij} = e_{ij} +\theta$, $e_{jk} =e_{jk}- \theta$, $e_{ki} =e_{ki} - \theta$\\
				$z_{ijk} =z_{ijk}- \theta$}
			}
			$\delta = $ sum of all changes in $e$ variables
		}
		{Define $\mb{E}$ with $\mb{E}(i,j) = e_{ij}, \forall i,j \in \{1, 2, \cdots, m\}$\\
		Return $\mb{D} = \mb{E} + \hat{\mb{D}}$}
	}
\end{algorithm}

\begin{table*}[t]
\caption{Evaluation results of comparison methods in learning graph distance. For the results that are not reported in the recent research works, the corresponding entries are marked with $-$ in the table. For the score of methods used to compute the ground truth, they are also provided for readers' reference, which are marked with * in the table.}\label{tab:main_result}
\centering
\small
\setlength{\tabcolsep}{5pt}
\renewcommand{\arraystretch}{1.0}
\begin{tabular}{|c|c|c|c||c|c|c||c|c|c|}
\hline
\multirow{3}{*}{\textbf{Methods}} & \multicolumn{9}{c|}{\textbf{Datasets}}\\
\cline{2-10}
&\multicolumn{3}{c||}{\textbf{AIDS}} &\multicolumn{3}{c||}{\textbf{LINUX}} &\multicolumn{3}{c|}{\textbf{IMDB}} \\
\cline{2-10}
&$\rho$	&$\tau$	&p@10	&$\rho$	&$\tau$	&p@10	&$\rho$	&$\tau$	&p@10	\\
\hline
\hline
GED* \cite{GED}	&1.000* &1.000* &1.000* &1.000* &1.000* &1.000* &NA &NA &NA\\
\hline
\hline
Beam \cite{Neuhaus_Fast_06}			&\textbf{0.609} &0.463 &\textbf{0.481} 	&\textbf{0.827} &\textbf{0.714} &\textbf{0.973} 	&- 	&0.837* &0.803* \\
VJ \cite{Fankhauser_Speeding_11}  		&\textbf{0.517} &0.383 &0.310  	&0.581 &0.450 &0.287  	&- 	&0.872* &0.825*  \\
Hungarian \cite{Riesen_Approximate_09} &0.510 &0.378 &0.360   	&\textbf{0.638} &0.517 &\textbf{0.913}  	&- 	&0.874* &0.815*\\
HED	\cite{Fischer_Approximation_15}	&- 	    &0.469 &\textbf{0.386}  	&- &\textbf{0.801} &\textbf{0.982} 		&-	 &0.627  &0.801 \\
\hline
\hline
EmbAvg \cite{Defferrard_Convolutional_16}									&-  &0.455 &0.176 	&- &0.012 &0.071 &- &0.179 &0.233\\
GCNMean \cite{Defferrard_Convolutional_16} 								&-  &\textbf{0.501} &0.186	&-   &0.424 &0.141 &- &0.307 &0.200\\
GCNMax \cite{Defferrard_Convolutional_16} 									&-  &\textbf{0.480 }&0.195 	&-   &0.495 &0.437 &- &\textbf{0.342} &\textbf{0.425}\\
Siamese MPNN \cite{8545310} 							&-  &0.210 &0.032 	&- &0.024 &0.009 &- &0.093 &0.023\\
\hline
\hline
{\our}	 								&0.551&0.440&0.226&0.613&0.534&0.332&\textbf{0.636}&\textbf{0.515}&\textbf{0.242}\\
{\our} (Triangle Fixing)						&\textbf{0.618}&\textbf{0.485}&\textbf{0.395}&\textbf{0.654}&\textbf{0.602}&0.516&\textbf{0.632}&\textbf{0.514}&\textbf{0.250}\\
\hline
\end{tabular}
\end{table*}

\begin{theorem}\label{theo:metric_optimization}
Given the collection of all potential distance matrices $\mc{D}$, the objective function $\mb{D}^* = \min_{\mb{D} \in \mc{D}} \left\| \mb{M} \odot (\mb{D} - \bar{\mb{D}}) \right\|_p$ can always attain its minimum on $\mc{D}$. Moreover, every local minimum is a global minimum. If, in addition, the norm is strictly convex and the weight matrix has no zeros or infinities off its diagonal, then there is a unique global minimum.
\end{theorem}
The main task to prove the theorem is to show that the objective function has no directions of recession, so it must attain a finite minimum on $\mc{D}$. Due to the limited space, we will not provide its proof here. If the readers are interested in the proof, you may also refer to  \cite{STD05} for more detailed information.

The main challenge in learning the framework lies in the constraints introduced by the \textit{triangle inequality} property on the distance metric, which will render the neural network very challenging to optimize. In this paper, we propose to train the model and obtain the final inferred distance matrix with two phases instead.

\noindent $\bullet$ \textbf{Step 1: Unconstrained Model Training}
Without considering the constraints, we can define the objective function in model learning as the following objective function:
\begin{equation}\label{equ:unconstrained}
\min \left\| \mb{M} \odot (\mb{D} - \bar{\mb{D}}) \right\|_p.
\end{equation}
Meanwhile, to lower down the learning cost, the {\gbert} component involved in {\our} can also be pre-trained in advance as discussed in Section~\ref{subsec:graph_bert}. Formally, based on the learned model, we can denote the inferred graph pairwise distance matrix as $\hat{\mb{D}} \in \mathbbm{R}^{m \times m}$, where the entries of the training instances are over-written with their true distance values.

\noindent $\bullet$  \textbf{Step 2: Constrained Metric Refining} 
Based on the real-number distance matrix, we can denote the constrained metric refining objective function as follows:
\begin{equation}
\hspace{-10pt}
\begin{aligned}
& \min \left\| \mb{D} - \hat{\mb{D}} \right\|_p \\
& s.t. \mb{D}(i,j) \hspace{-2pt} \le \hspace{-2pt} \mb{D}(i,k) \hspace{-2pt}+\hspace{-2pt} \mb{D}(j,k), \forall i, j, k \in \{1, \cdots, m\}.
\end{aligned}
\end{equation}
Distinct from Equations~\ref{equ:objective} and~\ref{equ:unconstrained}, term $\hat{\mb{D}}$ in the above equation is a constant matrix (not a variable any more). The only variable to be optimized in the above equation is $\mb{D}$, which actually can be reduced to the \textit{metric nearness} problem as studied in \cite{STD05}. In this paper, we will take the $L_2$ norm to define the loss function, and will use the triangle fixing algorithm as illustrated in Algorithm~\ref{alg:framwork} to help refine the learned distance metric values among the graph pairs. Formally, the output results of the triangle fixing algorithms will be returned as the final result.

\section{Experiments}\label{sec:experiment}

In this section, extensive experiments will be done on real-world benchmark datasets to test the effectiveness of {\our} proposed in this paper.


\subsection{Dataset and Experimental Settings}

\noindent \textbf{Dataset Descriptions}: The datasets used in this paper include AIDS \cite{Zeng_Comparing_09,Wang_An_12,Zhao_Partition_13,Zheng_Graph_13,Liang_Similarity_17}, LINUX \cite{Wang_An_12} and IMDB \cite{Yanardag_Deep_15}, which are all the benchmark datasets used in existing graph similarity search papers. For the AIDS and LINUX datasets (with small sized graphs), we will use the graph edit distance (GED) \cite{GED} as the ground truth; where as the ground truth of IMDB (graph instances in IMDB is much larger and their graph edit distance cannot be computed any more), we will compute the average of the graph pairwise Beam distance \cite{Neuhaus_Fast_06}, VJ distance \cite{Fankhauser_Speeding_11}, and Hungarian distance \cite{Riesen_Approximate_09} as the ground truth. To be more precise, the graph pairwise true distance between graph $G^{(i)}$ and $G^{(j)}$ is defined as $d(G^{(i)}, G^{(j)}) =1 - \exp(- \frac{d_{i,j}}{ (|\mc{V}^{(i)}| + |\mc{V}^{(j)}|)/2 }) \in [0, 1]$, where $d_{i,j}$ denotes the result computed by GED or the average of Beam, VJ and Hungarian as mentioned above.

\noindent \textbf{Experimental Settings}: For each dataset, the graph instances are partition into the train, validation and test sets according to the ratios: 6:2:2. Pairwise graph distance values will be computed for the graph instances in the training set; whereas the pairwise graphs between the those in the validation set and training set will be used to tune the model parameters. The final testing results are achieved for the pairwise graph instances between those from the testing set and the graphs in the whole dataset.

\noindent \textbf{Comparison Methods}: The comparison methods used in this paper include both classic combinatorial optimization based algorithms, e.g., Beam \cite{Neuhaus_Fast_06}, Volgenant-Jonker (VJ) \cite{Fankhauser_Speeding_11}, Hungarian \cite{Riesen_Approximate_09} and Hausdorff Edit Distance (HED) \cite{Fischer_Approximation_15}, and the recent deep neural network based graph distance algorithms, e.g., Message Passing Neural Networks (MPNN) \cite{8545310}, EmbAvg, GCN-Mean, and GCN-Max \cite{Defferrard_Convolutional_16}. The evaluation metrics adopted in this paper include \textit{Spearman's Rank Correlation Coefficient} ($\rho$), \textit{Kendall's Rank Correlation Coefficient} ($\tau$), and \textit{Precision@10}.

\noindent \textbf{Default Parameter Settings}: If not specified, the {\our} model used in this paper will have the following default parameter settings: \textit{input portal size}: $k=10$ (AIDS and LINUX) and $k=50$ (IMDB); \textit{hidden size}: 32; \textit{attention head number}: 2; \textit{hidden layer number}: $D=2$; \textit{learning rate}: 0.001; \textit{weight decay}: $5e^{-4}$; \textit{intermediate size}: 32; \textit{hidden dropout rate}: 0.5; \textit{attention dropout rate}: 0.3; \textit{graph residual term}: raw or none; \textit{training epoch}: 1000 (early stop when necessary).

\subsection{Experimental Results}

\noindent \textbf{Main Results}: The main results achieved by {\our} and the other baseline methods on the three benchmark datasets are provided in Table~\ref{tab:main_result}. Considering that GED and Beam/VJ/Hungarian are used to compute the ground truth on the AIDS/LINUX and IMDB, respectively, the evaluation scores obtained by these on the datasets (the scores are highlighted with *) are much better than the remaining methods. According to the table, since Beam is a fast approximated algorithm for computing the graph edit distance, its learning results are highly similar to the ground truth computed by the GED method on the AIDS and LINUX datasets. Meanwhile, among all the deep learning and graph neural network based distance metrics, {\our} and {\our} (Triangle Fixing) can out-perform them with great advantages. For these three studied datasets, {\our} and {\our} (Triangle Fixing) can rank among the top 3 for most of the evaluation metrics, which are highlighted in a bolded font in the table.

\noindent \textbf{With vs Without Triangle Inequality Constraint}: According to Table~\ref{tab:main_result}, for the {\our} method with the triangle inequality fixing process, it can improve the learning performance greatly on both AIDS and LINUX. Meanwhile, for the IMDB dataset, involving the triangle inequality fixing doesn't change the learning performance. Partial reason can be that instances in IMDB is relatively larger, which can provide more information to learn distinguishable representations. Further refining the learning results with the triangle fixing will not change most of the graph pairwise distance scores.

\begin{figure}[t]
    \centering
    	\includegraphics[width=0.95\linewidth]{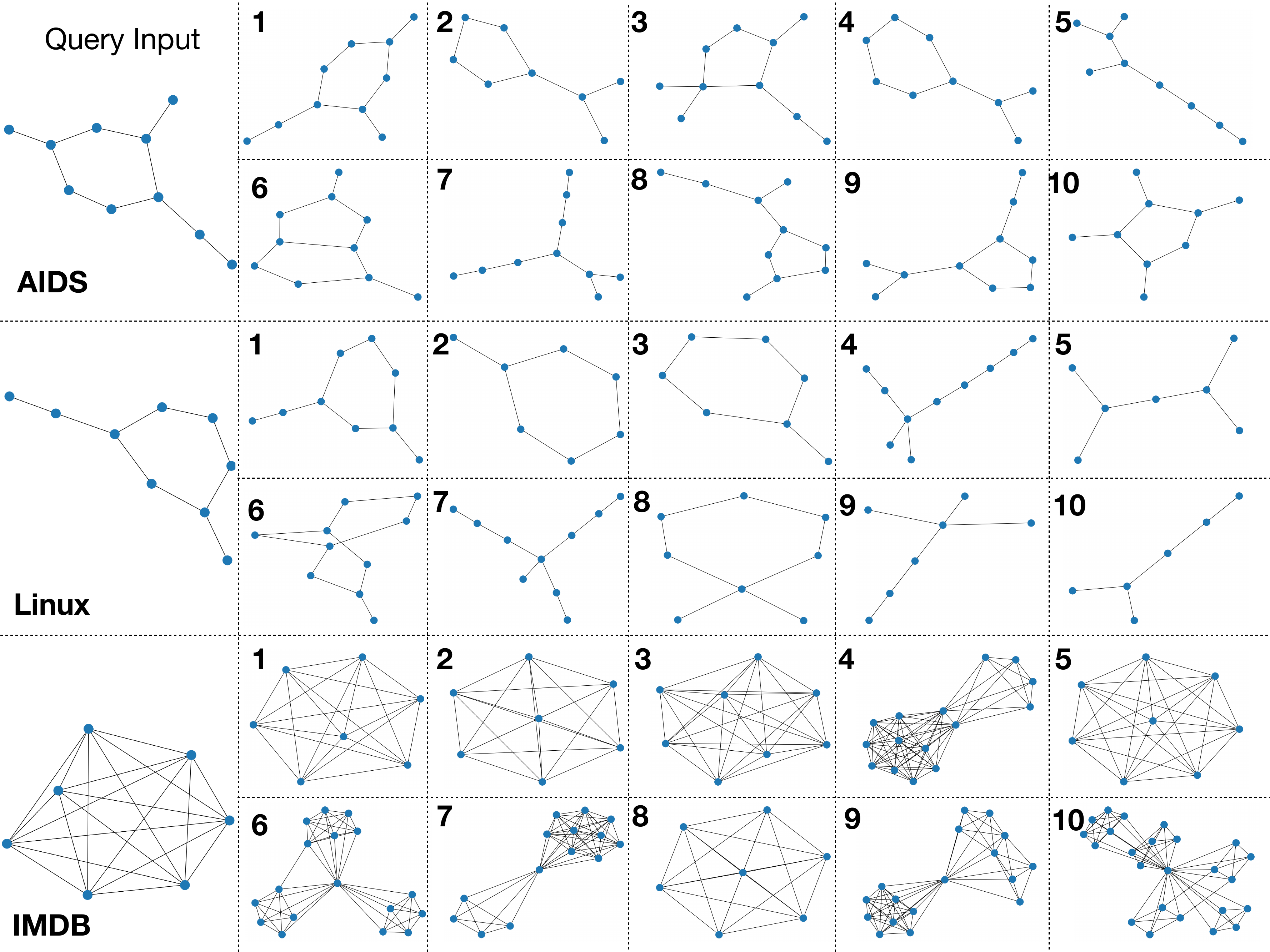}
    	\caption{Top 10 graph instances with the shortest distance to the query input graph (Row 1: AIDS, Row 2: Linux, Row 3: IMDB).}
    	\label{fig:case_study}
\end{figure}

\noindent \textbf{Case Studies}: As illustrated in Figure~\ref{fig:case_study}, we also show the top 10 graph instances in the datasets which has the shortest distance computed by {\our} to the input query graph. According to the results, {\our} can effectively identify the graph instances with similar structures to the input graph instance. For instance, the top 1 graph instance identified by {\our} on these three datasets all have the identical structure as the query graph input. Meanwhile, the remaining graph instances in the top 10 list also have very similar structures as the input graph.

\vspace{-5pt}
\section{Conclusion}\label{sec:conclusion}
\vspace{-5pt}

In this paper, we have studied the semi-supervised graph distance metric learning problem. To address the problem, a novel graph neural distance metric, i.e., {\our}, have been introduced. {\our} learns the graph instance representations by extending {\gbert} to the new problem settings, which can also be pre-trained in advance to lower-down the overall optimization time costs. In the learning process, the basic distance metric properties, i.e., \textit{non-negative}, \textit{identity of indiscernibles}, \textit{symmetry} and \textit{triangle inequality}, are all maintained, which can differentiate {\our} from most of the existing graph distance metric learning works. Extensive experiments done on real-world graph benchmark datasets also demonstrate the effectiveness of {\our} especially compared with the existing graph neural network based baseline methods.
\bibliography{reference}
\bibliographystyle{icml2020}


\end{document}